%% file: main.tex
\documentclass[11pt,twocolumn,letterpaper]{article}

\usepackage{cvpr}              
\usepackage{pgfplots}
\usepackage{pgfplotstable}
\usepackage{multirow}
\usepackage{caption}
\usepackage{float}
\usepackage{tablefootnote}
\usepackage{tcolorbox}
\usepackage[margin=1in]{geometry}    
\usepackage{graphicx}               
\usepackage{xcolor}                 
\usepackage[font=footnotesize]{caption}


\definecolor{cvprblue}{rgb}{0.21,0.49,0.74}
\usepackage[pagebackref,breaklinks,colorlinks,allcolors=cvprblue]{hyperref}

\def\paperID{58} 
\def\confName{CVPR}
\def\confYear{2025}

\title{Domain Adaptation of VLM for Soccer Video Understanding}

\author{
  Tiancheng Jiang\textsuperscript{1,2}\thanks{Work done during an internship at AWS.} \quad Henry Wang\textsuperscript{2} \quad Md Sirajus Salekin \textsuperscript{2} \\
  \quad Parmida Atighehchian \textsuperscript{2} \quad Shinan Zhang \textsuperscript{2} \\
  \textsuperscript{1}Massachusetts Institute of Technology
  \textsuperscript{2}Amazon Web Services \\
}

\begin{document}
\maketitle

\input{sec/0_abstract}    
\input{sec/1_intro}

\input{sec/2_related_work}
\input{sec/3_data_source_and_preprocessing}
\input{sec/4_methodology}
\input{sec/5_experiments}
\input{sec/6_conclusion}
{
    \small
    \bibliographystyle{ieeenat_fullname}
    \bibliography{main}
}


\end{document}



\include{suppl}

%% file: sec/0_abstract.tex
\begin{abstract}
 Vision Language Models (VLMs) have demonstrated strong performance in multi-modal tasks by effectively aligning visual and textual representations. However, most video understanding VLM research has been domain-agnostic, leaving the understanding of their transfer learning capability to specialized domains underexplored. In this work, we address this by exploring the adaptability of open-source VLMs to specific domains, and focusing on soccer as an initial case study. Our approach uses large-scale soccer datasets and LLM to create instruction-following data, and use them to iteratively fine-tune the general-domain VLM in a curriculum learning fashion (first teaching the model key soccer concepts to then question answering tasks). The final adapted model, trained using a curated dataset of 20k video clips, exhibits significant improvement in soccer-specific tasks compared to the base model, with a 37.5\% relative improvement for the visual question-answering task and an accuracy improvement from 11.8\% to 63.5\% for the downstream soccer action classification task. 
\end{abstract}

%% file: sec/1_intro.tex
\section{Introduction}
\label{sec:intro}
Recent advances in multimodal large language models and vision language models (VLMs) have significantly improved their ability to process and understand visual inputs, including images and videos \cite{Bordes2024AnIT}. However, while general-domain video VLMs have been extensively studied, there remains a gap in understanding how well these models can be adapted to domain-specific applications \cite{jiang2024gpt4vdistribution, huang2024vtimellm, fang2024mmbenchvideo}. This paper aims to address this gap by exploring an effective methodology and repeatable recipe for adapting a general-purpose video VLM to a specialized domain. Furthermore, since fine-tuning video VLMs tends to be computationally expensive, we wanted to explore an approach that's not only simple and practical but efficient at the same time.

As a case study, we investigate the adaptation of LLaVA-NeXT-Video \cite{zhang2024llavanextvideo}, a state-of-the-art open-source video VLM, to the soccer domain in sports. Soccer presents unique challenges for video understanding due to its dynamic nature, rapid motion of objects, and fine-grained events \cite{giancola2018soccernet, deliege2021soccernet}.  
To this end, we employ a structured three-stage fine-tuning approach using synthetically generated data. First, the model undergoes \textbf{Concept Alignment}, where it is trained to associate soccer-specific concepts with video clips to improve visual-text alignment. Next, \textbf{Instruction Tuning} enhances the model’s ability to answer soccer-specific queries. Finally, \textbf{Downstream Task Fine-tuning} aims to boost model performance on specific soccer-related tasks. This methodology ensures a gradual and effective adaptation of the model to the target domain. We evaluate the performance of the adapted model on multiple tasks, including caption generation, visual question answering, and action classification. 
\begin{figure*}[t]
    \centering
    \includegraphics[width=\textwidth]{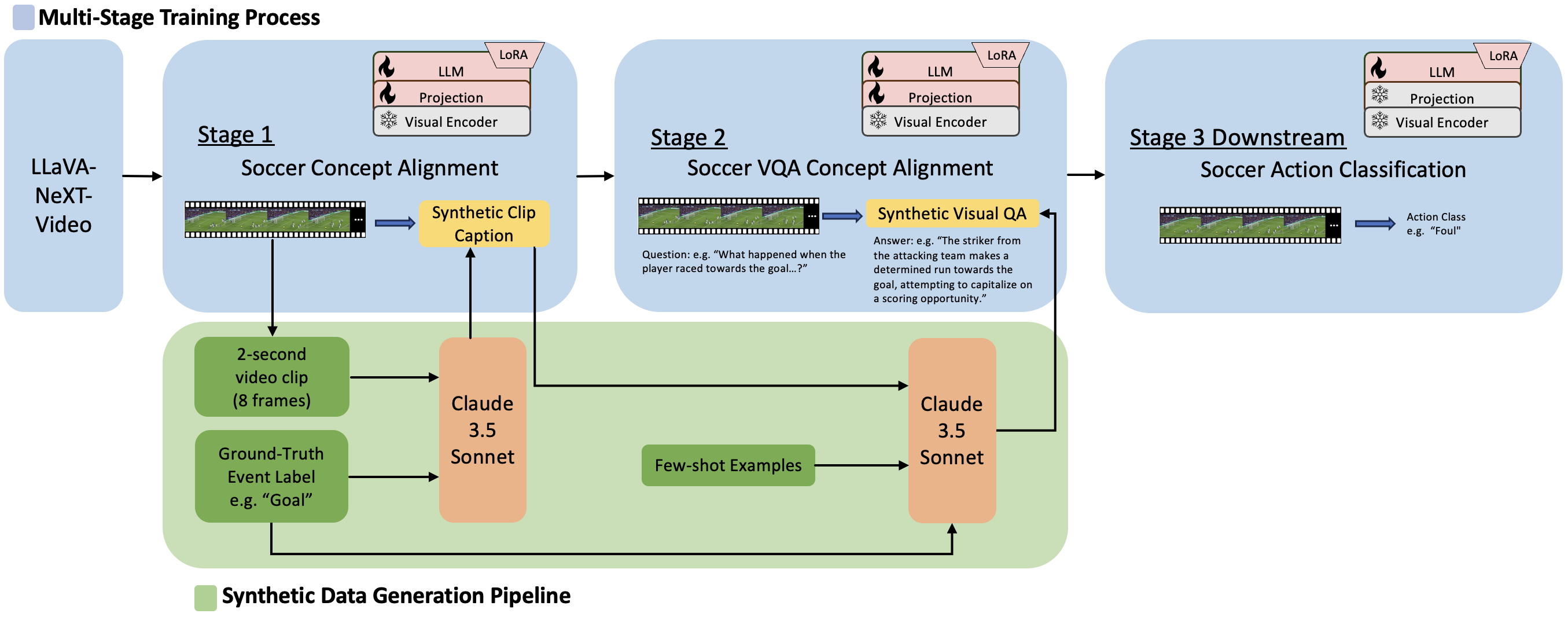} 
    \caption{Overall workflow of our proposed multi-stage domain adaptation process. The LLaVA-Next-Video model is iteratively trained in a curriculum learning fashion (first teaching the model key soccer concepts to then question answering tasks). During the first two training stages, visual encoder is frozen while training both the projector and LLM modules using LoRA. For downstream task, only the LLM modules are trained with LoRA. The figure also demonstrates the synthetic data generation pipeline instruction tuning data that is generated for the first two training stages using Claude 3.5 Sonnet V1.} 
    \label{fig:multi-stage_training_data_gen}
\end{figure*}

Our contributions are summarized below: 
\begin{itemize}
    \item We introduce a simple, yet effective multi-stage fine-tuning strategy for adapting video VLMs to specialized domains. We further demonstrate the success of our proposed adaptation strategy through the lens of a real-world case study in soccer, where a general-purpose video VLM learns to effectively analyze different tasks in this challenging sports domain.
    \item We formalize best practices through various ablation studies, providing insight on the specific design choices of the training process for domain adaptation.
\end{itemize}

%% file: sec/2_related_work.tex
\section{Related work}
\label{sec:related_work}
\textbf{The Rise of VLM.}
Building on the success of large language models (LLMs), vision language models (VLMs), which integrate vision encoders with LLMs for multi-modal reasoning, have seen a tremendous amount of improvements in performance that span across tasks like image captioning, segmentation, classification, etc. over the last few years \cite{li2019visualbert, lu2019vilbert, radford2021learning, li2023blip2, liu2024llava}. Multi-stage training that first aligns vision and language representation followed by instruction tuning appears to emerge as an effective approach \cite{lin2023vila}. VLMs have also been extended for challenging video understanding tasks, which requires temporal modeling. To this date, however, most of the work has been focused on creating general-purpose video VLMs. \cite{zellers2021merlot, zhang2023videollama, ataallah2024minigpt4video, lin2023video, Maaz2023VideoChatGPT}

\noindent\textbf{VLM Domain Adaptation.}
Domain adaptation of VLMs has been an active research area. Med-Flamingo \cite{moor2023medflamingo} and LLaVA-Med \cite{li2023llavamed}, have demonstrated effective adaptation to the medical domain by fine-tuning general-purpose VLMs on biomedical data. Other studies have extended into fields such as robotics \cite{rt22023arxiv}, scientific literature \cite{li2025chainofregion}, remote sensing \cite{silva2024large}, etc. For example, RT-2 \cite{rt22023arxiv} adapted a vision-language model for robotics by training it on web-scale multimodal data alongside robotic demonstrations, allowing it to translate visual understanding into actionable robotic tasks. Despite these advances, video VLM domain adaptation remains under explored as most efforts have been focused on image-based tasks. The scarcity of high-quality labeled datasets complicates domain adaptation for video-based VLMs.

\noindent\textbf{Sports Understanding.}
Sports understanding \cite{thomas2017computervision} has been an evolving field. It includes tasks such as, action recognition \cite{deliege2021soccernet, giancola2018soccernet, gu2020basketball}, commentary generation \cite{qi2019sports, rao2024matchtime, xi2024knowledge, yu2018finegrained}, intelligent refereeing \cite{Held2024XVARS-arxiv, held2023vars}, etc. With the rise of multi-modal large language models (MLLMs) and VLMs, there have also been works focused at creating generalized frameworks for multi-sports understanding \cite{li2024sportsqa, xia2024sportqa}. Soccer analysis, in particular, traditionally was focused on action spotting \cite{deliege2021soccernet, mkhallati2023soccernetcaption}, replay grounding \cite{held2023vars, zhou2021baidusoccer}, player tracking \cite{cioppa2022soccernettracking, vandeghen2022soccerdetection}, and foul recognition \cite{Held2024XVARS-arxiv, held2023vars}, and had largely been facilitated by the SoccerNet datasets \cite{cioppa2022soccernettracking, deliege2021soccernetv2, gautam2024soccernetechoes, giancola2018soccernet}. As an initial case study of our multi-stage training process, we explore adapting a general-purpose video VLM to the soccer domain. 

%% file: sec/3_data_source_and_preprocessing.tex
\section{Video Data Source and Preprocessing}
\label{sec:data}

We source videos from two large-scale soccer datasets, SoccerNet-V2 \cite{deliege2021soccernet} and WyScout \cite{wyscout}, both of which provide timestamped key soccer event labels. 

\subsection{Video Sources}
SoccerNet-V2 consists of 550 full-length broadcast matches from six major European leagues, with Action Spotting labels marking 17 key soccer events, such as ``Goal'', ``Throw-in'', etc. WyScout, a subscription-based proprietary soccer dataset, includes over 2,000 games from different leagues worldwide and offer event labels including both primary and secondary actions. It covers a broader range of in-play events like passes, duels, and tactical actions. Given the differences in soccer event labels, the two datasets capture distinct though overlapping distributions of soccer activities, making their combined use beneficial for model training.

For both datasets, we extract 2-second clips based on event timestamps while preserving their respective training-test splits (300/100 games for SoccerNet; a 4:1 ratio for WyScout). These clips are then used to create instruction-following data for model adaptation. 

\subsection{Preprocessing}
\textbf{Video Clips}.
To prepare the video clips for fine-tuning, we process full soccer matches into 2-seconds long clips. Each 2-second long game is represented by uniformly selecting 8 frames, a common choice for many open-source video VLMs \cite{lin2023video, Maaz2023VideoChatGPT} and allows the context to be continuous. As each clip corresponding to one key soccer event, it starts 0.5 seconds before and ends 1.5 seconds after the event's timestamp to capture both the lead-up and aftermath of a soccer event. The rationale behind the 2-seconds duration can be found in \cref{sec:experiments:ablations}. 

\noindent\textbf{Soccer Event Distribution}. Once the clips were generated from the event labels, we randomly sample to ensure uniform distribution of soccer events in both the training and test set. In SoccerNet, however, due to the super rare nature of 4 event categories (Red card, Yellow \textrightarrow Red card, Penalty, Kick off), we decided to remove them so the resulting dataset contains clips representing 13 soccer events instead of the original 17. 

%% file: sec/4_methodology.tex
\section{Methodology}
\label{sec:method}

In this section, we introduce LLaVA-NeXT-Video, a general-domain video VLM, as the base VLM used in this research. We then provide an overview of our multi-stage domain adaptation process.

\subsection{Base Model and Architecture}

LLaVA-Next-Video \cite{zhang2024llavanextvideo}, a state-of-the-art open-source general video VLM, extends the capabilities of LLaVA-Next (LLaVA-1.6) \cite{liu2024llava} by integrating video-specific processing enhancements while maintaining strong performance in multimodal reasoning tasks. 

LLaVA-Next-Video \cite{zhang2024llavanextvideo} incorporates a \textit{CLIP-ViT} visual encoder, which maps image and video frames into latent feature representations. A lightweight MLP projection layer connects the visual encoder to the language model, ensuring efficient multi-modal integration. The model also applies two architectural changes to enhance video comprehension, which are AnyRes and Length Generalization. The model is pretrained on a mix of multi-modal datasets, including large-scale image-text and video-text data, making it a strong base model to build upon. The same network architecture is used in this research. To fine-tune the model, we create instruction following dataset and train the model using its original auto-regressive training objective. We calculate the probability of the target response \( X_r \) as the following. 
\begin{equation}
    p(X_r | X_v, X_q) = \prod_{i=1}^{L} p_{\theta} \left( x_i \mid X_v, X_{q, <i}, X_{r, <i} \right)
\end{equation}
where \( X_v \) and \( X_q \) indicate the video clip and the instruction or question respectively. \(\theta \) is the trainable parameters and \(X_{q, <i}\) and \(X_{r, <i}\) are the instruction and answer tokens before the current prediction token \( X_i \).

\subsection{Multi-Stage Fine-tuning Process}

Our proposed fine-tuning process as shown in \Cref{fig:multi-stage_training_data_gen} consists of three training stages: 1) Soccer Concept Alignment, 2) Soccer Visual QA Instruction Tuning, and 3) Downstream Task Fine-tuning. Each stage progressively improves the model's ability to align and understand the soccer domain. For efficient fine-tuning, we employ the Low Rank Adaptation (LoRA) method \cite{hu2022lora} in all three training stages. The full training was completed using four 24G A10 GPUs.  

\noindent\textbf{Stage 1 - Soccer Concept Alignment}. 
The first training stage, Soccer Concept Alignment, teaches the base model fundamental soccer concepts by associating video clips with their corresponding event labels. We generate a synthetic clip-caption dataset by prompting Claude 3.5 Sonnet V1 \cite{claude35sonnet} to describe each 2-second clip based on its ground-truth event label and the 8 extracted frames. This approach ensures accurate and contextually grounded captions, reducing hallucinations that arise when generating descriptions purely from visual inputs.

From the clip-caption pairs, we then apply a structured expansion method to convert them into instruction-following samples: each sample consists of a video clip and an instruction prompting the model to describe the action, with the corresponding caption as the target response. We manually curate a set of synonymous instructions (e.g. ``Please provide a description of what happened in the soccer match video?'') and randomly assign one per sample to create some variability in the instruction.

During training, we only freeze the visual encoder and apply LoRA adapters to the language model, and train using the model's original task on next-token prediction on the ground-truth captions. 

\noindent\textbf{Stage 2 - Soccer VQA Instruction Tuning}. 
The second training stage, Soccer VQA Instruction Tuning, aims to train the model to respond to diverse soccer-related queries. We construct this stage's instruction-following dataset using the synthetically generated captions from Stage 1, prompting Claude 3.5 Sonnet V1 to generate five distinct question-answer pairs per caption:
\begin{itemize}
    \item \textbf{Description Questions:} Asking for a textual description of the video.
    \item \textbf{Temporal Questions:} Querying the sequence of events to test time-based reasoning.
    \item \textbf{Causal Questions:} Understanding cause-effect relationships between actions.
    \item \textbf{Prediction Questions:} Asking what might happen next based on visual context.
    \item \textbf{Action Spotting Questions:} Identifying key soccer actions in the clip.
\end{itemize}
To maintain dataset balance, we randomly sample from each question type, ensuring equal distribution while keeping dataset size consistent with Stage 1. We also enforce a 4:1 ratio of free-form to multiple-choice formats.
During training, we also only freeze the visual encoder and apply LoRA adapters to the language model and train using the model's original next-token prediction task on the ground-truth query answers. 

\noindent\textbf{Stage 3 - Downstream Task Fine-tuning}. 
The final stage fine-tunes the model on any downstream task in the domain. The purpose of this stage's training is mainly to teach the model the specific formatting needed for the downstream task. 

\noindent\textbf{Instruction Following Dataset Details}. 
All instruction-following datasets consist of a 2-second long soccer video clip \( X_v \), an instruction or question \( X_q \), and an associated response \( X_r \), all following the same format: 
\begin{quote}
    \texttt{Human: \(X_v\) \(X_q\)<STOP>\textbackslash n} \\
    \texttt{Assistant: \(X_r\)<STOP>\textbackslash n}
\end{quote}
For the first two training stages, we create three dataset sizes, 3k, 10k, and 20k samples, with the smaller datasets as subsets of the larger ones, to understand the effect of dataset sizes on model performance. Also, to ensure privacy and generalization, we prompt Claude 3.5 Sonnet V1 to anonymize player and team names when generating synthetic data.

%% file: sec/5_experiments.tex
\section{Experiments}
\label{sec:experiments}
\subsection{Evaluation Tasks and Metrics}

We designed three evaluation tasks.

\noindent\textbf{Task 1: Caption Generation}. This task evaluates the model’s ability to generate accurate and descriptive captions for 2-sec soccer clips. Two test sets were curated: a 1,000-sample set and a 200-sample subset for ablation studies. Unlike the training data in Stages 1 and 2, where the 3k and 10k datasets only contain SoccerNet clips, both test sets here contain an equal split between SoccerNet and WyScout samples. Captions were generated using the same approach as in Stage 1.

We employed two sets of evaluation metrics. The first was LLM-as-a-Judge metrics, where we asked Claude 3.5 Sonnet V1 to rate the predicted captions against the label captions, and two scores from 1 to 5 (5 being the best) were provided for Correctness and Detailness. The average for either criteria was computed. The second set of evaluation metrics included traditional captioning metrics such as BLEU and ROUGE scores.

\noindent\textbf{Task 2: Visual Question Answering (VQA)}. This task assessed the model’s ability to answer soccer-related queries. Two test sets (1,000 and 200 samples) were also constructed for the same reason, and both maintain a balanced distribution across the five question types and a 4:1 ratio of free-form to multiple-choice formats. Each test set included an equal split of clips from SoccerNet and WyScout.

For evaluation, we used Claude 3.5 Sonnet V1 to score both the predicted response and a reference upper bound response on helpfulness, relevance, accuracy, and level of details from 1 to 10 (10 being the highest score). We compute a relative score by normalizing the predicted response score with the upper bound response score, and an average relative score for the whole test set is calculated. A higher score indicates better alignment with the reference and a score exceeding 100 suggests the model outperformed the reference in certain cases.

\noindent\textbf{Task 3: SoccerNet Action Classification}. This task was chosen as an example downstream task, and it evaluated the model’s ability to classify 2-sec soccer clip into one of 13 predefined soccer action classes in an instruction following format (no classifier head was added). The training set consists of 3.3k samples, which is small as the model is expected to have already acquired soccer domain knowledge from the earlier training stages. The test set consisted of two sizes, a larger set of 1,300 samples (containing 100 samples per event class) and a smaller test set if 100 samples (containing approximately even distribution across classes). The performance was measured using standard classification metrics including accuracy, precision, recall, and F1-score. 

\subsection{Experimental Results}
For the main experiments, we adapted the base Llava-NeXT-Video model through the first two stages of the fine-tuning process using different sizes of training sets, and used the 3.3k training set for the third fine-tuning stage on Action Classification as the downstream task. We report results of the three evaluation tasks below.  

\noindent\textbf{Caption Generation Task}. We used the model after Stage 2 for this task. The results from \cref{tab:capgen_results} indicate that the 20k model (model trained using the 20k samples dataset) consistently outperforms the other models across both correctness and detailness scores. However, interesting trends can be observed when analyzing the performance of the 3k and 10k models, which were trained only on SoccerNet clips. For example, the 3k model’s average detailness or correctness scores decreased in comparison to that of the Base model, even though SoccerNet-specific scores improved. This decline was attributed to lower WyScout-specific scores, highlighting limitations in the 3k model’s ability to generalize to unseen distributions in WyScout. The 10k model also displayed lower WyScout-specific scores than the Base model, but it overall performed better than the 3k model, indicating perhaps some generalization effects. The inclusion of WyScout data in the training set for the 20k model resulted in significant performance gains across both subsets of the test set, demonstrating the value of a mixed training set. Traditional metrics, such as BLEU-4 and ROGUE in Tab. 6 of the Appendix, showed a large improvement from the Base model to the 3k model, but subsequent increases with larger datasets were less pronounced.
\begin{table*}[t]
    \centering
    \small
    \captionsetup{justification=centering, width=\textwidth} 
    \begin{tabular}{lccc|ccc}
        \toprule
        & \multicolumn{3}{c|}{\textbf{Correctness Scores}} & \multicolumn{3}{c}{\textbf{Detailness Scores}} \\
        & \textbf{Total Avg.} & \textbf{SoccerNet } & \textbf{WyScout} & \textbf{Total Avg.} & \textbf{SoccerNet} & \textbf{WyScout} \\
        (Question Count) & (1000) & (500) & (500) & (1000) & (500) & (500)\\
        \midrule
        Claude 3.5 Sonnet V1 & \textbf{2.779} & \textbf{2.594} & \textbf{2.964} & 2.402 & 2.420 & 2.384 \\
        LLaMA 3.2\footnotemark[1] & 1.875 & 1.748 & 2.002 & 2.159 & 2.166 & 2.159 \\
        Base Model & 2.130 & 2.158 & 2.102 & 2.226 & 2.276 & 2.176 \\
        \midrule
        \multicolumn{7}{l}{\textbf{Adapted Models}} \\
        3K (SoccerNet clips only)  & 1.782 & 2.214 & 1.350 & 2.048 & 2.436 & 1.660 \\
        10K (SoccerNet clips only) & 1.905 & 2.314 & 1.496 & 2.2 & 2.566 & 1.834 \\
        20K & 2.528 & 2.446 & 2.610 & \textbf{2.675} & \textbf{2.616} & \textbf{2.734} \\
        \bottomrule
    \end{tabular}
    \caption{Performance comparison of Caption Generation task measured by LLM as Judge scored on Correctness and Detailness.}
    \label{tab:capgen_results}
\end{table*}

\noindent\textbf{Visual Question Answering Task}.
For VQA, we also used the model after Stage 2 for this task. From \cref{fig:vqa_graph} and \cref{tab:vqa_table}, we can see that the 20k model achieved the highest relative scores, reflecting the strongest performance. There was also a steady improvement from models trained using smaller to larger datasets, with the largest gains between the Base and 3k models and between the 10k and 20k models. The sharp improvement between the 10k and 20k models not only signals the importance of having more data but it also emphasize the importance of incorporating WyScout data in the training set as it broadens the data distribution. Looking at specific question types, the largest gains were observed in Action Recognition, Description, and Temporal question types. 
\begin{figure}
    \centering
    \small
    \begin{tikzpicture}
        \begin{axis}[
            ybar,
            bar width=8pt,
            symbolic x coords={Base Model, 3k Model, 10k Model, 20k Model},
            xtick=data,
            ymin=0,
            ymax=100,
            ymajorgrids=true,
            xticklabel style={rotate=45, anchor=east},
            enlarge x limits=0.2, 
            legend style={at={(0.5,1.15)}, anchor=north, legend columns=-1} 
        ]
        \addplot coordinates {(Base Model, 60.20) (3k Model, 71.31) (10k Model, 73.77) (20k Model, 82.79)};
        \addplot coordinates {(Base Model, 57.44) (3k Model, 78.96) (10k Model, 82.77) (20k Model, 83.76)};
        \addplot coordinates {(Base Model, 62.97) (3k Model, 63.67) (10k Model, 64.81) (20k Model, 81.83)};
        
        \legend{Relative Score, SoccerNet RS, WyScout RS}
        
        \end{axis}
    \end{tikzpicture}
    \captionsetup{justification=raggedright, singlelinecheck=false}
    \caption{Comparison of the VQA task abilities measured by the relative score.}
    \label{fig:vqa_graph}
\end{figure}
\begin{table*}[t]
    \centering
    \footnotesize
    \captionsetup{justification=centering, width=\textwidth} 
    \label{tab:performance}
    \begin{tabular}{lcc|ccccc|c}
        \toprule
        & \multicolumn{2}{c|}{\textbf{Question Sources}} & \multicolumn{5}{c|}{\textbf{Domains}} & \textbf{Overall} \\
        & \textbf{SoccerNet} & \textbf{WyScout} & \textbf{Desc.} & \textbf{Temp.} & \textbf{Causal} & \textbf{Pred.} & \textbf{Action Rec.} \\
        (Question Count) & (500) & (500) & (250) & (250) & (250) & (250) & (250) & (1000) \\
        \midrule
        Claude 3.5 Sonnet V1 & 75.79 & 74.85 & 42.72 & \textbf{91.74} & 83.57 & 95.50 & 63.07 & 75.32 \\
        LLaMA 3.2\footnotemark[1] & 69.97 & 85.66 & 40.60 & 67.42 & \textbf{97.38} & \textbf{126.77} & 56.92 & 77.82 \\
        Base Model & 57.44 & 62.97 & 36.21 & 61.49 & 83.49 & 79.26 & 40.57 & 60.20 \\
        \midrule
        \multicolumn{9}{l}{\textbf{Adapted Models}} \\
        3K (SoccerNet clips only) & 78.96 & 63.67 & 47.62 & 69.91 & 83.84 & 89.95 & 65.25 & 71.31 \\
        10K (SoccerNet clips only) & 82.77 & 64.81 & 51.31 & 72.49 & 82.41 & 93.58 & 69.07 & 73.77 \\
        20K & \textbf{83.76} & \textbf{81.83} & \textbf{59.14} & 81.91 & 91.07 & 97.85 & \textbf{84.02} & \textbf{82.79} \\
        \bottomrule
    \end{tabular}
    \captionsetup{justification=raggedright, singlelinecheck=false}
    \caption{Performance comparison of VQA task abilities, measured by the relative score via Claude 3.5 Sonnet V1 evaluation, normalized against a reference response. A higher score indicates better alignment with the reference and a score exceeding 100 suggests the model outperformed the reference in certain cases.}
    \label{tab:vqa_table}
\end{table*}
\footnotetext[1]{\fontsize{8}{9}\selectfont Only a single frame selected at timestamp of the event was used as input for the model due to API limitation. We tested using the 90B version of LLaMA 3.2.}
\begin{figure*}[t]
    \centering
    \includegraphics[width=\textwidth]{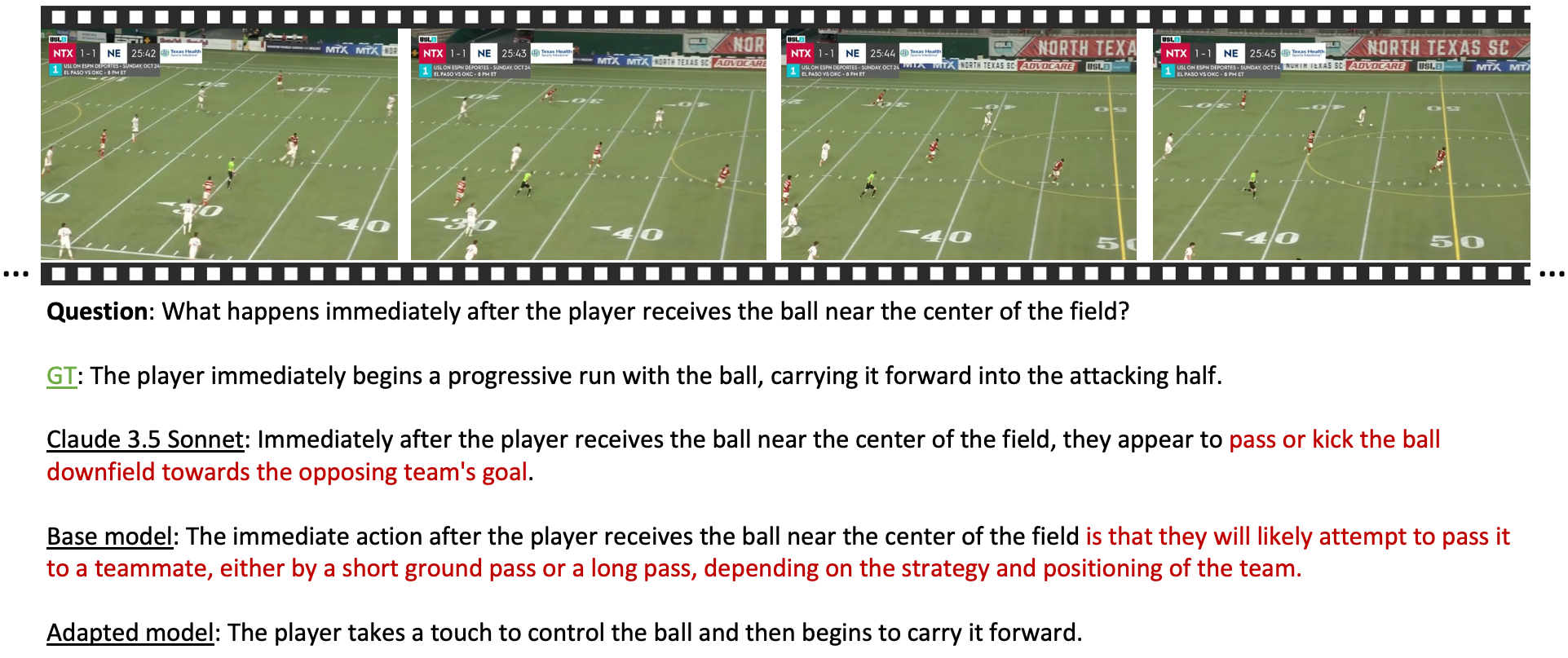} 
    \caption{A qualitative example from the VQA task test set showing the ground-truth answer and responses from the different models. The red indicates hallucination}
\end{figure*}

\begin{figure*}[t]


    \begin{subfigure}{0.48\textwidth}
        \centering
        \includegraphics[width=0.85\linewidth]{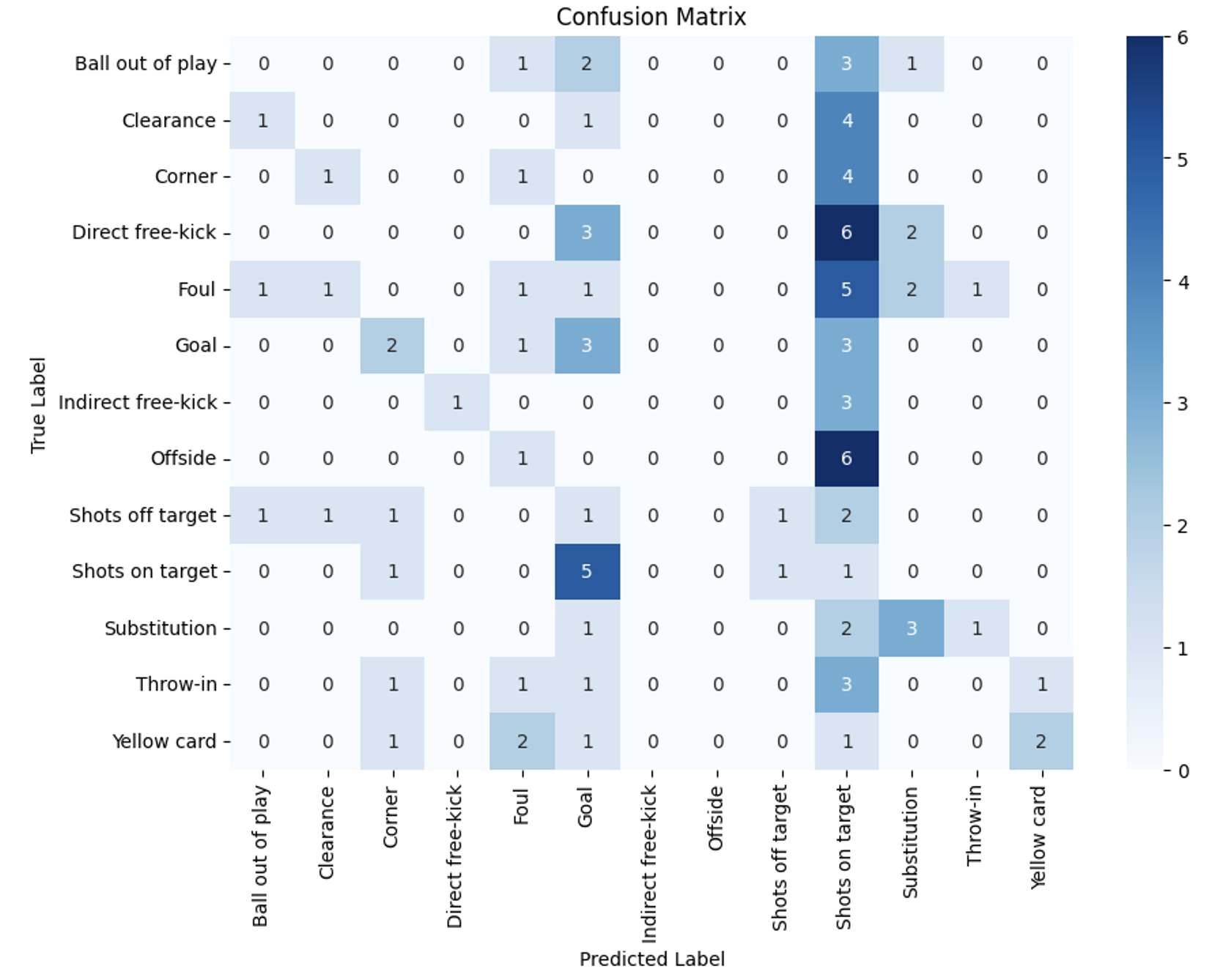}
        \caption{Base Model Pred.}
        \label{fig:base-model}
    \end{subfigure}
    \hfill
    \begin{subfigure}{0.48\textwidth}
        \centering
        \includegraphics[width=0.85\linewidth]{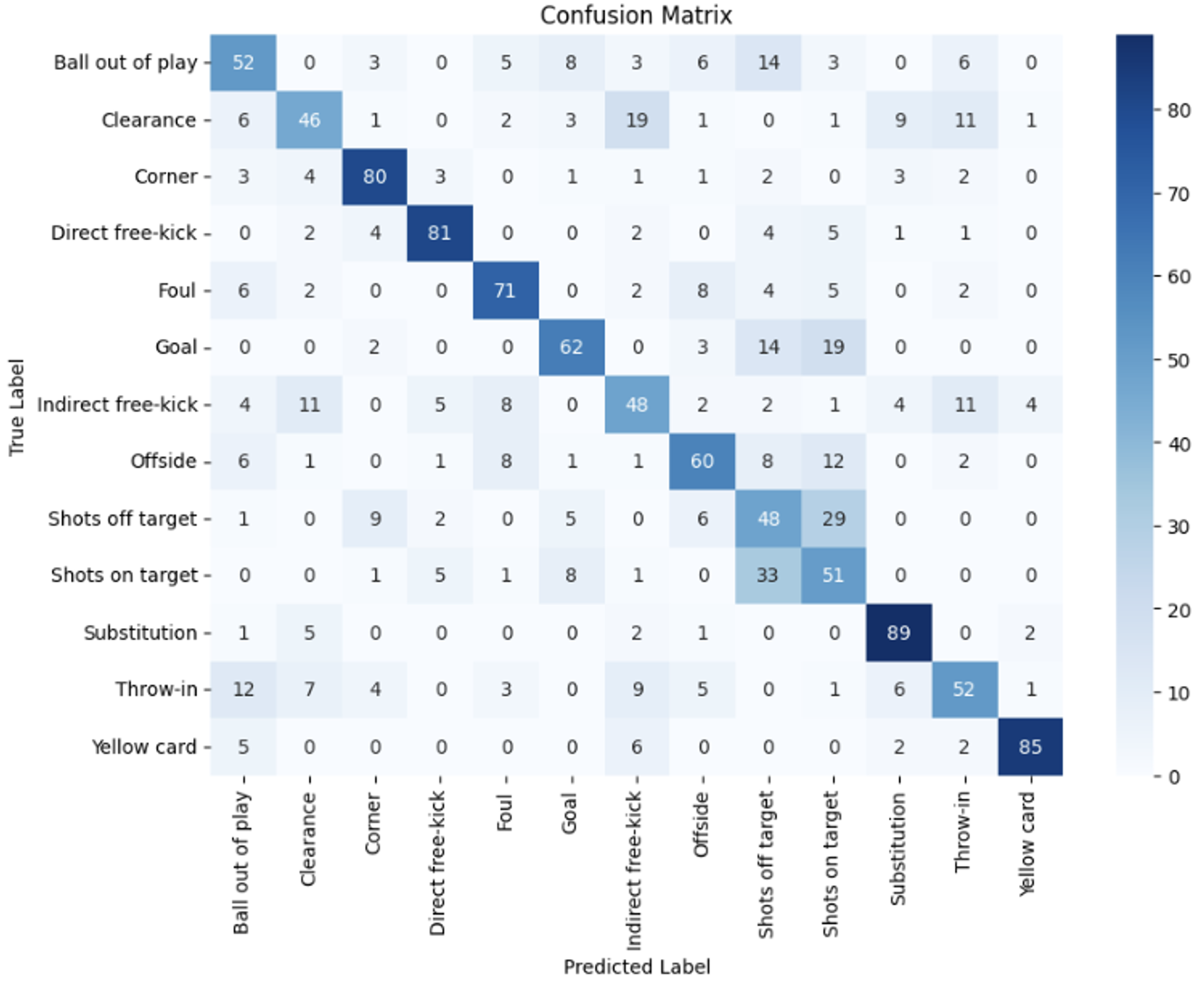}
        \caption{\textbf{20k Model Pred.}}
        \label{fig:20k-model}
    \end{subfigure}

    \caption{Confusion Matrices from the Action Classification task for the Base model (LLaVA-NeXT-Video), and our adapted 20k Model. Their testset prediction accuracies are 0.118, and 0.635 respectively. Full results can be found in Tab. 7 of the Appendix}
    \label{fig:ac_confusion}
\end{figure*}

\noindent\textbf{Soccer Action Classification Task}.
For Soccer Action Classification, since it's a downstream task, we used the model at the end of stage 3 to run the test set. From \cref{fig:ac_confusion}, one can see that the accuracy improves significantly from the Base model to that of the 20k model, increasing from 11.8\% to 57.8\%. However, the confusion matrix of the 20k model prediction revealed that the model struggles to distinguish between visually similar actions, such as “Shots on Target” and “Shots off Target,” suggesting that additional training strategies may be required to refine fine-grained classification. As benchmarks, LLaMA 3.2 and Claude 3.5 Sonnet V1 scored an accuracy of 24.2\% and 26.7\% respectively. 
    
\subsection{Ablation Studies and Best Practices}
\label{sec:experiments:ablations}
\noindent\textbf{Training the Projector or Not}. We compared training only the LLM versus training both the LLM and the projector through the first two stages of the training process. Based on results shown in Tab. 8 and 9 in the appendix, we suggest training on both components as it provides incremental benefits without substantial computational overhead.

\begin{figure*}[t]
    \centering
    \includegraphics[width=\textwidth]{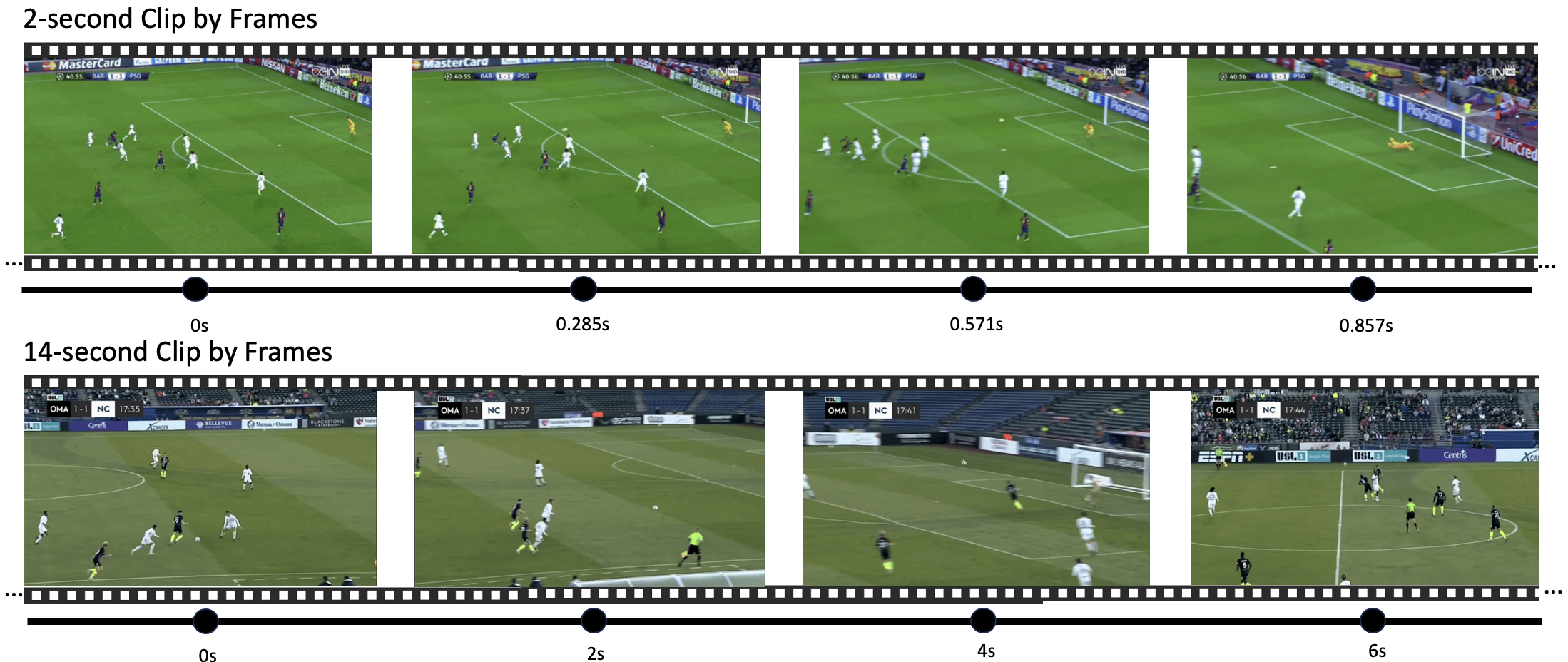} 
    \caption{Demonstration of short clip vs. long clip on information continuity. The above 2s clip is much more continuous hence easier to comprehend while the bottom 14s clip has more information loss due to the wide time gap between the frames.}
    \label{fig:long_short_clip}
\end{figure*}

\noindent\textbf{Rationale for 2-Second Clip Duration}.
To adapt video VLMs for soccer analysis, we segmented full matches into 2-sec clips. While longer clips are typically used as pre-training data, we found that for soccer domain such duration can introduce excessive noise as it captures overlapping soccer events and increases hallucination risks in synthetic data generation. In contrast, 2-sec clips provide a focused context, ensuring a strong association with the soccer event label. This effectively reduces ambiguity and enhances data alignment. Additionally, with a fixed eight-frame constraint, shorter clips minimize information loss in this fast-paced domain. \cref{fig:long_short_clip} illustrates how shorter clips offer clearer event representation, making them more interpretable even for humans.

\noindent\textbf{Multi-Stage Training Process}. Our results highlighted the importance of the multi-stage training approach in both VQA performance and action classification accuracy. As shown in \cref{tab:vqa_procedure}, while Instruction Tuning (IT) provided the largest boost in VQA performance, the best results were achieved when the model underwent both Concept Alignment (CA) and IT, indicating that CA establishes essential soccer knowledge before instruction tuning. Similarly, Tab.14 in Appendix demonstrates that directly fine-tuning the base model on action classification training set yields minimal improvement, whereas models trained through the full CA → IT → Action Classification (AC) sequence achieved the highest accuracy. 
\begin{table}[t]
    \centering
    \captionsetup{justification=centering, width=\columnwidth} 
    \label{tab:performance}
    \begin{tabular}{lc}
        \toprule
        \textbf{Visual QA Task} & \textbf{RS} \\
        \midrule
        Base Model  & 58.57 \\
        20k - Base$\rightarrow$Concept Alignment(CA) & 54.85 \\
        20k - Base$\rightarrow$Instruction Tuning(IT) & 79.09 \\
        20k - Base$\rightarrow$CA$\rightarrow$IT & \textbf{85.80} \\
        \bottomrule
    \end{tabular}
    \captionsetup{justification=raggedright, singlelinecheck=false}
    \caption{Comparison of performance for the VQA task across different training sequence variants, measured by the average relative score. }
    \label{tab:vqa_procedure}
\end{table}

\noindent\textbf{Extension to longer duration clips} The fine-tuned model, trained on 2-sec clips, performed similarly when classifying 5-sec clips, achieving a macro F1 score of 0.61 compared to 0.63 on 2-sec clips. This suggests early evidence of the model's generalizability to varying clip lengths, which could improve efficiency in downstream tasks.

\subsection{Limitations} While our training relies on synthetic data, which may introduce stylistic biases, it provides a strong foundation for developing and validating our methodology. Our data pipeline, though effective, can be prone to hallucination for longer clips. We evaluated our approach using LLaVA-Next-Video but have not yet tested other video VLMs, particularly those with longer context windows, which presents opportunities for future explorations. We also did not fine-tune the visual encoder, which could further enhance the model’s ability to capture fine-grained, domain-specific visual cues.



%% file: sec/6_conclusion.tex
\section{Conclusion}
\label{sec:conclusions}
This work demonstrates the feasibility of adapting a general-purpose video VLM to a specialized domain through a structured multi-stage fine-tuning approach using synthetically generated instruction-following data. Looking ahead, we aim to extend our evaluation to additional downstream tasks beyond soccer action classification to more real-world soccer applications to further validate the generalizability of our approach. Additionally, we plan to apply this approach in other specialized domains to assess and refine the broader generalizability of our multi-stage fine-tuning procedure. 

%% file: suppl.tex
\clearpage
\setcounter{page}{12}

\textbf{\huge Appendix}  
\vspace{10pt}  

\section{Data}
\textbf{Instructions for video clip caption.} 
The list of instructions used to instruct the model to describe or caption the soccer activity shown in the video clip. They present the same meaning with natural language variance. All samples from Stage's Concept Alignment's training set use one of the following instructions.

\vspace{0.5cm} 

\setcounter{table}{4}
\begin{tcolorbox}[
    colframe=black, colback=lightgray, 
    sharp corners=all, boxrule=0.5pt, 
    width=\textwidth
]
    \begin{itemize}
        \item ``Please provide a description of what happened in the soccer match video?''
        \item ``What key events took place in this soccer video clip?''
        \item ``Describe breifly what happened in this soccer game video?''
        \item ``Provide a short description of the video.''
        \item ``Give an account of the activites captured in the video.''
        \item ``What is a quick summary of the soccer events in the video?''
        \item ``Present a compact description of the video clip.''
        \item ``Relay a brief, clear account of the soccer video clip shown.''
    \end{itemize}
\end{tcolorbox}

\captionsetup{justification=centering}
\captionof{table}{The list of instructions for video clip description.}

\setcounter{figure}{6}
\begin{figure}[H]
    \centering
    \begin{tcolorbox}[
        colframe=black, colback=lightgray, 
        sharp corners=south, rounded corners=north,
        title={\textbf{Instruction for Soccer Action Classification Task}},
        fonttitle=\bfseries,
        width=\textwidth, 
        colbacktitle=black, 
        coltitle=white
    ]
    \small 
    \ttfamily 
    \begin{verbatim}
Identify the key action that happened in the given video clip of a soccer game.

Select one from the list below: 
- Ball out of play 
- Clearance 
- Corner 
- Direct free-kick 
- Foul 
- Goal 
- Indirect free-kick 
- Offside 
- Shots off target
- Shots on target 
- Substitution
- Throw-in 
- Yellow card 

Note: Only answer with one action from the list and nothing else.
    \end{verbatim}
    \end{tcolorbox}

    \caption{The same instruction is applied to all samples in the training and test set for the Soccer Action Classification Task. The order of the list of actions is randomized for each sample.}
\end{figure}
\clearpage
\section{Prompt}
\begin{figure}[H]
    \centering
    \begin{tcolorbox}[
        colframe=black, colback=lightgray, 
        sharp corners=south, rounded corners=north,
        title={\textbf{Prompt to Generate Synthetic Captions}},
        fonttitle=\bfseries,
        width=\textwidth, 
        colbacktitle=black, 
        coltitle=white
    ]
    \footnotesize 
    \ttfamily 
    \begin{verbatim}
# Character
You are an AI assistant specialized in soccer topics and in anlayzing broadcast soccer 
games. You are provided with 8 sequential frames of a 2 seconds long broadcast soccer 
match video clip that represents one soccer event. You are provided with the 
groundtruth annotation of the soccer event that took place during the clip. Using the 
frames and the key soccer action groundtruth annotation, please describe or caption 
what you see in the soccer match clip. Focus response on the keep soccer event of the 
groundtruth label, and only add visual information when appropriate. 

# Requirements and Helpful Info
Below are requirements for generating the description of the video clip:
    - Respond as if you have watched the clip instead of just select frames. 
    - DO NOT make things up. Base your response only on the video frames and the 
    groundtruth annotation.
    - Anonymize the team names and refer to them as team A and B, but stay consistent 
    in the description. DO NOT mention the league name as well. 
    - DO NOT mention jersey number and player name. 
    - DO NOT mention jersey color. 
    - You MUST use each frame's groundtruth annotation in your response but answer 
    as if you were only given the broadcast soccer match frames. 
    - DO NOT use phrases like "frame", "mentioned", "caption", "context" in the 
    response. 
    - Use the visual information in the video frame to make sense of the groundtruth 
    annotations. 
    - DO NOT start the response using 'In this broadcast soccer match' or similar.
    Answer directly with the actual description.
    - Keep your response concise.
    - Focus response on the keep soccer event of the groundtruth label.
    
# Input Format 
Broadcast Soccer Match Frame 1: <frame1>
Broadcast Soccer Match Frame 2: <frame2>
Broadcast Soccer Match Frame 3: <frame3>
Broadcast Soccer Match Frame 4: <frame4>
Broadcast Soccer Match Frame 5: <frame5>
Broadcast Soccer Match Frame 6: <frame6>
Broadcast Soccer Match Frame 7: <frame7>
Broadcast Soccer Match Frame 8: <frame8>

Key Soccer Event in the Clip: <event>

# Output Format
Put your response in the <clip_description></clip_description> tag.
    \end{verbatim}
    \end{tcolorbox}
    \captionsetup{justification=raggedright, singlelinecheck=false}
    \caption{We used the above to prompt Claude 3.5 Sonnet to generate synthetic captions for the 2-second clips, which are then put in instruction following format for training in Concept Alignment.}
\end{figure}

\begin{figure}[t]
    \centering
    \begin{tcolorbox}[
        colframe=black, colback=lightgray, 
        sharp corners=south, rounded corners=north,
        title={\textbf{Prompt to Generate Synthetic Question Answer Pairs}},
        fonttitle=\bfseries,
        width=\textwidth, 
        colbacktitle=black, 
        coltitle=white
    ]
    \small 
    \ttfamily 
    \begin{verbatim}
# Task 
Given a detailed description and groundtruth soccer event that summarize the content 
of a soccer game video clip, generate question-answer pairs. The question-answer 
pairs should be faithful to the content of the video description and developed from 
different dimensions to promote comprehensive understanding of the video. Here are 
some question dimensions and their explanations and exampled question-answer pairs 
for reference: 
{question_definitions}

# Guidelines For Question-Answer Pairs Generation:
- Read the video description provided carefully, paying attention to the content, 
such as the scene where the video takes place, the main characters and their 
behaviors, and the development of the key events.
- Generate appropriate question-answer pairs based on the description. The 
question-answer pairs should cover as many question dimensions and not deviate 
from the content of the video description.
- Generate one question-answer pair for each question dimension.
- Multiple choice question should include multiple answer options as part of the 
question. The answer must contain just one answer choice. 
- Generate answer to question as if you have watched the clip and not the description. 
- Generate question-answer pairs in such as a way that the answer cannot be derived 
based on common sense from the question and requires understanding of the video. 
- Generate most of the questions as free form answers but the rest as multiple choice
questions as shown in the examples. 4 questions free form, 1 question multiple choice.
Only 'Description' questions must have free-form answer, and other types of questions
could all be multiple choice questions.  
- Team A and B are anonymized and interchangeable so don't make questions and answers
specific to team references. Make it generic. 
- Use different multiple choice question and answer formats as shown in the examples. 
For example, use 'A, B,...' to reference answer choices and sometimes use just '-'
to separate the choices. Also for example, sometimes the answer should just be the
answer choice, but sometimes use a full sentence that mentions the answer choice. 
- Aside 'Prediction' types questions, all other question types should base their
questions and answers off of the description and ground truth soccer event(s).   

# Input Format 
Description: {caption}, 
Key Soccer Event: {event}

# Output Format:
1. Your output should be formed in a JSON file.
2. Only provide the Python dictionary string.
Your response should look like and no other text:
[{"Dimension": <dimension-1>, "Question": <question-1>, "Answer":<answer-1>, 
"Is_multiple_choices": <True_or_False>},
{"Dimension": <dimension-2>, "Question": <question-2>, "Answer":<answer-2>, 
"Is_multiple_choices": <True_or_False>},...] 
    \end{verbatim}
    \end{tcolorbox}
    \captionsetup{justification=raggedright, singlelinecheck=false}
    \caption{We used the above to prompt Claude 3.5 Sonnet to generate synthetic question-answer pairs for the different question types for training in Instruction Tuning.}
\end{figure}

\begin{figure}[t]
    \centering
    \begin{tcolorbox}[
        colframe=black, colback=lightgray, 
        sharp corners=south, rounded corners=north,
        title={\textbf{Question Type Definitions and Few-Shot Examples}},
        fonttitle=\bfseries,
        width=\textwidth, 
        colbacktitle=black, 
        coltitle=white
    ]
    \footnotesize 
    \ttfamily 
    \begin{verbatim}
# Task 1 - Description: this task is designed to assess the ability of the model in generating 
an informative description of the video clip. It would essentially be rephrasing the provided 
caption. 
## caption-1: A player from team A, wearing the number 11 jersey, is being substituted off the
field. He hugs and celebrates with his teammates and coaching staff on the sidelines before 
exiting the pitch. The player appears joyful and is all smiles as he embraces his colleagues, 
likely after a positive contribution to the match.
## quesiton-1: Please provide a description of what happened in the soccer match video?
## answer-1: A player from Team A, sporting the number 11 jersey, is leaving the field as a 
substitute. He shares hugs and celebrations with his teammates and coaches on the sidelines 
before stepping off the pitch. The player looks delighted, beaming as he greets his colleagues, 
likely in recognition of a strong performance during the game.
...

# Task 2 - Temporal: this task is designed to assess the model's capability of reasoning the 
temporal order of events in the video clip, for example what activity happened before another. 
## caption-1: The goalkeeper of team A takes an indirect free kick from inside the penalty area. 
He kicks the ball forward towards the center of the field as the match continues.
## question-1: What happened right before the goalkeeper kicked the ball towards the center of
the field?
## answer-1: The goalkeeper took an indirect free kick from inside the penalty area, then 
kicked the ball forward as the match continued.
...

# Task 3 - Causal: this task is designed to assess the model's ability to understand 
the causality between events. 
## caption-1: Team A takes a corner kick. The ball is played into the penalty area towards a
group of players from both teams jostling for position. Team B manages to clear the ball away
from danger as the corner kick fails to produce a scoring opportunity.
## question-1: How did the ball end up in the penalty area?
## answer-1: The ball ended up in the penalty area because Team A took a corner kick and aimed 
it toward the group of players near the goal.
...

# Task 4 - Prediction: this task is designed to assess the model's ability to understand what's 
going in the video clip and based of it make the right prediction of what might happen next. 
## caption-1: Team A player commits a foul on a team B player in the middle of the field. 
The referee blows the whistle to stop play and award a free kick to team B. 
## question-1: What is likely to happen next after the recent interaction between the players 
in the clip?
## answer-1: The referee will likely stop play and award a free kick to the opposing team, 
Team B, because the clip shows a player from Team A committing a foul on a Team B player in
the middle of the field. According to soccer rules, a foul in open play typically results in
the opposing team gaining a free kick, allowing them to restart play from the spot of the
infraction.
... 

# Task 5 - Action Recognition: this task is designed to assess the model's ability to identify 
the key soccer action(s) or event(s) (such as Goal, Foul, etc) that took place in the short 
video clip.  
## caption-1: Team A takes a corner kick. The ball is played into the penalty area towards a 
group of players from both teams jostling for position. Team B manages to clear the ball away
from danger as the corner kick fails to produce a scoring opportunity.
## question-1: What key soccer event or activity took place in the video clip? 
## answer-1: The key soccer event shown in the clip is a corner kick where Team A sent the ball
into the penalty area where players from both teams were competing for position. 
...
    \end{verbatim}
    \end{tcolorbox}
    \captionsetup{justification=raggedright, singlelinecheck=false}
    \caption{Definition of the five question types and some few-shot examples embedded in Figure 3's prompt to generate synthetic question answer pairs.}
\end{figure}

\clearpage

\section{Additional Main Experiments Results}
\begin{table}[H]
    \centering
    \footnotesize
    \begin{tabular}{lcccccc}
        \toprule
        \textbf{Metric} & \textbf{Claude 3.5 Sonnet} & \textbf{LLaMA 3.2} & \textbf{Base Model} & \textbf{3K Model} & \textbf{10K Model} & \textbf{20K Model} \\
        \midrule
        BLEU 1 & 0.154 & \textbf{0.312} & 0.289 & 0.287 & 0.290 & 0.288 \\
        BLEU 4 & 0.019 & 0.029 & 0.026 & \textbf{0.088} & \textbf{0.088} & 0.082 \\
        ROUGE & 0.158 & 0.189 & 0.208 & \textbf{0.264} & \textbf{0.264} & 0.260 \\
        \bottomrule
    \end{tabular}
    
    \caption{Traditional metrics results for Caption Generation task across models.}
    \label{tab:model-comparison}
\end{table}
\begin{table}[H]
    \centering
    \scriptsize
    \begin{tabular}{lcccccc}
        \toprule
        \textbf{Metric} & \textbf{LLaMA 3.2} & \textbf{Claude 3.5 Sonnet} & \textbf{Base Model} & \textbf{3K Model} & \textbf{10K Model} & \textbf{20K Model} \\
        \midrule
        Accuracy & 0.242 & 0.267 & 0.118 & 0.578 & 0.623 & \textbf{0.635} \\
        Macro Precision & 0.268 & 0.283 & 0.114 & 0.589 & 0.628 & \textbf{0.644} \\
        Macro Recall & 0.245 & 0.266 & 0.118 & 0.578 & 0.623 & \textbf{0.635} \\
        Macro F1 & 0.199 & 0.253 & 0.092 & 0.578 & 0.623 & \textbf{0.637} \\
        Weighted Precision & 0.238 & 0.283 & 0.114 & 0.589 & 0.628 & \textbf{0.644} \\
        Weighted Recall & 0.242 & 0.267 & 0.118 & 0.578 & 0.623 & \textbf{0.635} \\
        Weighted F1 & 0.172 & 0.253 & 0.092 & 0.579 & 0.623 & \textbf{0.637} \\
        \bottomrule
    \end{tabular}
    
    \caption{Performance comparison for the Soccer Action Classification task across different models.}
    \label{tab:model-comparison}
\end{table}

\clearpage
\section{Ablation Studies Results}
\subsection{Training the Projector or Not}
\begin{table}[H]
    \centering
    \small
    \captionsetup{justification=centering, width=\textwidth} 
    \label{tab:performance}
    \begin{tabular}{lccc|ccc}
        \toprule
        & \multicolumn{3}{c|}{\textbf{Correctness Scores}} & \multicolumn{3}{c}{\textbf{Detailness Scores}} \\
        & \textbf{Avg.} & \textbf{SoccerNet } & \textbf{WyScout} & \textbf{Avg.} & \textbf{SoccerNet} & \textbf{WyScout} \\
        (Question Count) & (200) & (100) & (100) & (200) & (100) & (100)\\
        \midrule
        \multicolumn{7}{l}{\textbf{Adapted Models}} \\
        3K - LLM Only  & 1.76 & 2.11 & 1.40 & 1.955 & 2.29 & 1.62 \\
        3K - Projector+LLM & \textbf{1.92} & \textbf{2.43} & \textbf{1.41} & \textbf{2.155} & \textbf{2.55} & \textbf{1.76} \\
        \bottomrule
    \end{tabular}
    \captionsetup{justification=raggedright, singlelinecheck=false}
    \caption{Results comparison for the Caption Generation task between the two model variants in this ablation study}
\end{table}
\begin{table}[H]
    \centering
    \captionsetup{justification=centering, width=\textwidth} 
    \label{tab:performance}
    \begin{tabular}{lc|cc}
        \toprule
        & \textbf{Overall} & \multicolumn{2}{c}{\textbf{Question Sources}} \\
        & & \textbf{SoccerNet} & \textbf{WyScout} \\
        (Question Count) & (200) & (100) & (100) \\
        \midrule
        \multicolumn{4}{l}{\textbf{Adapted Models}} \\
        3K - LLM Only  & 71.85 & 79.356 & 66.35 \\
        3K - Projector+LLM & \textbf{72.99} & \textbf{80.89} & \textbf{65.09} \\
        \bottomrule
    \end{tabular}
    \captionsetup{justification=raggedright, singlelinecheck=false}
    \caption{Results comparison for the Visual QA task between the two model variants in this ablation study}
\end{table}

\clearpage
\subsection{Impact of LoRA Rank Sizes}
We investigated the impact of different LoRA rank sizes by training models using Rank 32 and Rank 64. The results show very little differences and mixed results between the Caption Generation and VQA tasks. Given this, we only opted for Rank 64 when scaling up training to the 20k datasets, as it provides greater capacity for handling complex patterns.
\begin{table}[H]
    \centering
    \small
    \captionsetup{justification=centering, width=\textwidth} 
    \label{tab:performance}
    \begin{tabular}{lccc|ccc}
        \toprule
        & \multicolumn{3}{c|}{\textbf{Correctness Scores}} & \multicolumn{3}{c}{\textbf{Detailness Scores}} \\
        & \textbf{Avg.} & \textbf{SoccerNet } & \textbf{WyScout} & \textbf{Avg.} & \textbf{SoccerNet} & \textbf{WyScout} \\
        (Question Count) & (200) & (100) & (100) & (200) & (100) & (100)\\
        \midrule
        \multicolumn{7}{l}{\textbf{Adapted Models}} \\
        10K - 32 Rank  & 2.08 & 2.63 & 1.53 & 2.38 & 2.84 & 1.91 \\
        10K - 64 Rank & \textbf{2.16} & \textbf{2.67} & \textbf{1.64} & \textbf{2.48} & \textbf{2.91} & \textbf{2.04} \\
        \bottomrule
    \end{tabular}
    \captionsetup{justification=raggedright, singlelinecheck=false}
    \caption{Results comparison for the Caption Generation task between the two model variants in this ablation study.}
\end{table}
\begin{table}[H]
    \centering
    \small
    \captionsetup{justification=centering, width=\textwidth} 
    \label{tab:performance}
    \begin{tabular}{lc|cc}
        \toprule
        & \textbf{Overall} & \multicolumn{2}{c}{\textbf{Question Sources}} \\
        & & \textbf{SoccerNet} & \textbf{WyScout} \\
        (Question Count) & (200) & (100) & (100) \\
        \midrule
        \multicolumn{4}{l}{\textbf{Adapted Models}} \\
        10K - 32 Rank  & \textbf{78.15} & 86.59 & \textbf{69.22} \\
        10K - 64 Rank & 77.94 & \textbf{87.36} & 68.53 \\
        \bottomrule
    \end{tabular}
    \captionsetup{justification=raggedright, singlelinecheck=false}
    \caption{Results comparison for the Visual QA task between the two model variants in this ablation study.}
\end{table}

\clearpage
\subsection{Effect of Incorporating Images in Training for Concept Alignment}
Another ablation study examined whether incorporating image data alongside video data in the Concept Alignment training stage improves performance or not. While including images and training sequentially with images first then video clips produced slight gains in performance across tasks, when we increased the quantity of image data or the number of epochs it led to incoherent outputs. As a result, we excluded image data from the main experiments. However, testing training with both images and videos data may be a interesting direction for future researches to improve model performance.
\begin{table}[H]
    \centering
    \small
    \captionsetup{justification=centering, width=\textwidth} 
    \label{tab:performance}
    \begin{tabular}{lccc|ccc}
        \toprule
        & \multicolumn{3}{c|}{\textbf{Correctness Scores}} & \multicolumn{3}{c}{\textbf{Detailness Scores}} \\
        & \textbf{Avg.} & \textbf{SoccerNet } & \textbf{WyScout} & \textbf{Avg.} & \textbf{SoccerNet} & \textbf{WyScout} \\
        (Question Count) & (200) & (100) & (100) & (200) & (100) & (100)\\
        \midrule
        \multicolumn{7}{l}{\textbf{Adapted Models}} \\
        10K - No Image  & 1.9 & 2.29 & 1.51 & 2.22 & 2.56 & 1.87 \\
        10K - Image then Video & \textbf{2.08} & \textbf{2.63} & \textbf{1.53} & \textbf{2.38} & \textbf{2.84} & \textbf{1.91} \\
        10K - image and Video Mixed & 1.85 & 2.11 & 1.58 & 2.19 & 2.48 & 1.89 \\
        \bottomrule
    \end{tabular}
    \captionsetup{justification=raggedright, singlelinecheck=false}
    \caption{Results comparison for the Caption Generation task between the three model variants in this ablation study.}
\end{table}
\begin{table}[H]
    \centering
    \captionsetup{justification=centering, width=\textwidth} 
    \label{tab:performance}
    \begin{tabular}{lc|cc}
        \toprule
        & \textbf{Overall} & \multicolumn{2}{c}{\textbf{Question Sources}} \\
        & & \textbf{SoccerNet} & \textbf{WyScout} \\
        (Question Count) & (200) & (100) & (100) \\
        \midrule
        \multicolumn{4}{l}{\textbf{Adapted Models}} \\
        10K - No Image  & 78.15 & 86.59 & 69.22 \\
        10K - Image then Video & \textbf{79.04} & 87.30 & \textbf{70.78} \\
        10K - image and Video Mixed & 76.18 & \textbf{87.48} & 64.88 \\
        \bottomrule
    \end{tabular}
    \captionsetup{justification=raggedright, singlelinecheck=false}
    \caption{Results comparison for the Visual QA task between the three model variants in this ablation study.}
\end{table}

\subsection{The Necessity of the Multi-Stage Training Process}
    

\begin{table}[H]
    \centering
    \captionsetup{justification=centering, width=\columnwidth} 
    \label{tab:performance}
    \begin{tabular}{lc}
        \toprule
        \textbf{Action Classification (AC)} & \textbf{Accuracy} \\
        \midrule
        Base Model  & 11.8\% \\
        Base$\rightarrow$AC  & 16\% \\
        20k - Base$\rightarrow$Concept Alignment(CA) & 23\% \\
        20k - Base$\rightarrow$CA$\rightarrow$AC & 52\% \\
        20k - Base$\rightarrow$CA$\rightarrow$IT$\rightarrow$AC & \textbf{63.5\%} \\
        \bottomrule
    \end{tabular}
    \captionsetup{justification=raggedright, singlelinecheck=false}
    \caption{Results comparison for the Action Classification task between the five model variants in this ablation study. It illustrates the values of the different training stages as the model variant that undergoes the full process achieves the best result. }
    \label{tab:vqa_procedure}
\end{table}

\clearpage

\subsection{Some More Qualitative Examples}
\begin{figure}[H]
    \centering
    \includegraphics[width=\textwidth]{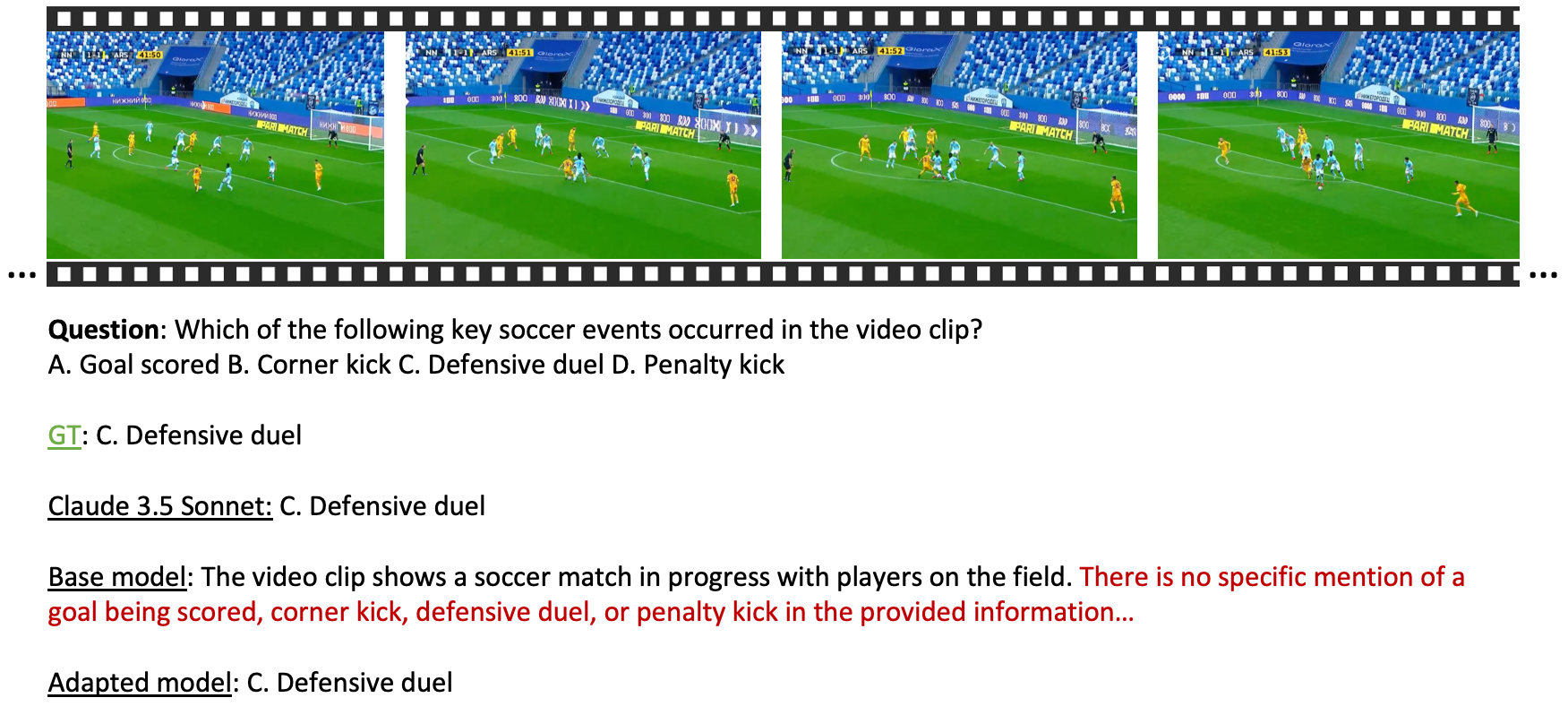} 
    \captionsetup{justification=raggedright, singlelinecheck=false}
    \caption{A multiple choice example from the VQA task test set showing the ground-truth answer and responses from the different models. The red indicates hallucination.}
    \label{fig:full-width}
\end{figure}
\begin{figure}[H]
    \centering
    \includegraphics[width=\textwidth]{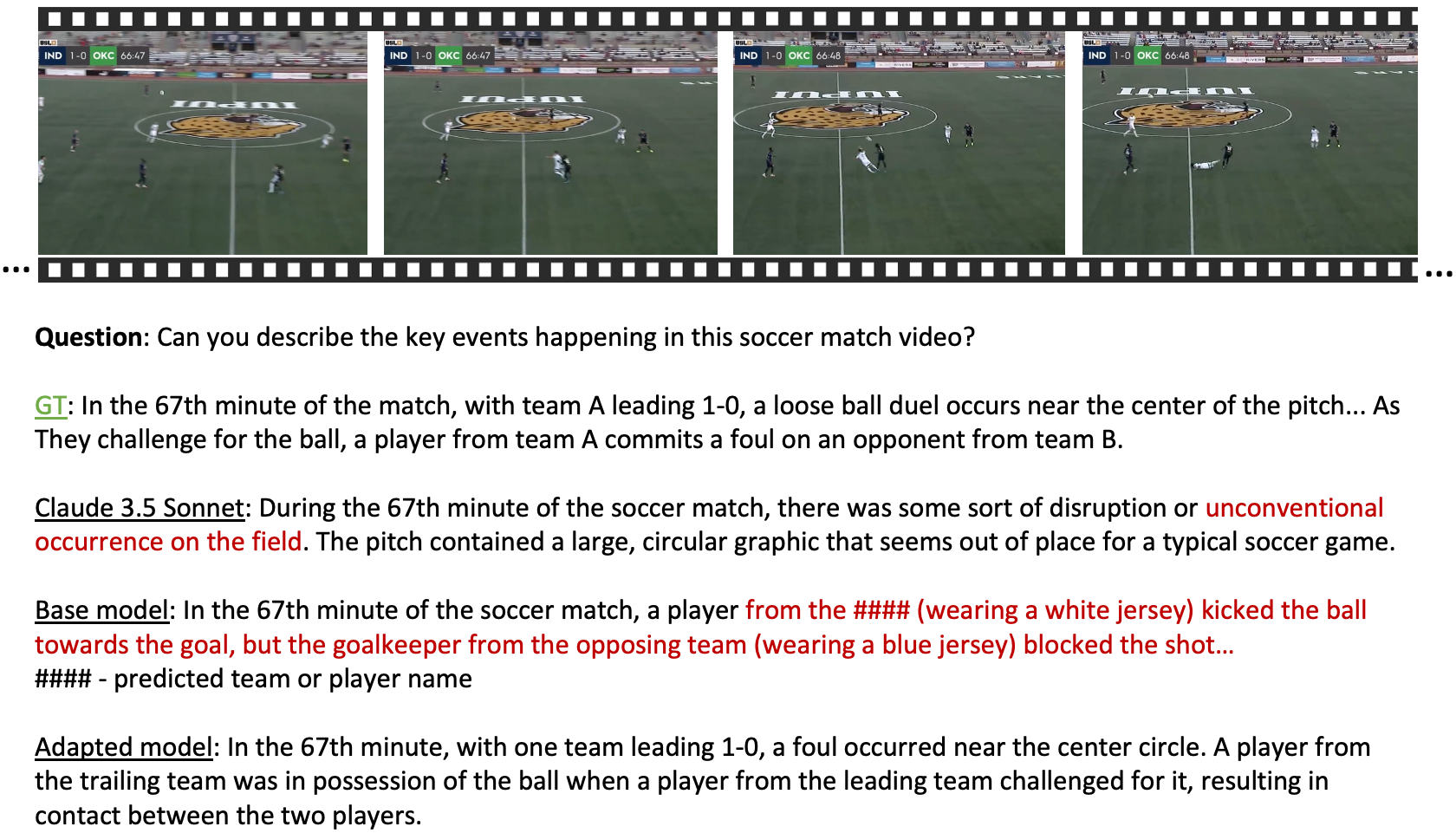} 
    \captionsetup{justification=raggedright, singlelinecheck=false}
    \caption{A free response example from the VQA task test set showing the ground-truth answer and responses from the different models. The red indicates hallucination.}
    \label{fig:full-width}
\end{figure}